\documentclass{article}
\usepackage{spconf,amsmath,graphicx}
\usepackage{amssymb}
\usepackage{booktabs}
\usepackage{pifont}
\usepackage{multirow}
\usepackage{cancel}
\usepackage{tikz}
\usetikzlibrary{positioning}
\usetikzlibrary{calc}
\usepackage{pgfplots, pgfplotstable}
\pgfplotsset{compat=1.17} 
\usepackage{url}

\newcommand{\Sref}[1]{\S\ref{#1}}
\newcommand{\Fref}[1]{Figure~\ref{#1}}
\newcommand{\cmark}{\ding{51}}%
\newcommand{\Tref}[1]{Table~\ref{#1}}

\newcommand{\Hquad}{\hspace{0.0em}} 
\newcommand\mypar[1]{\noindent\textbf{#1}\Hquad}

\usepackage{icomma}

\makeatletter
\def\blfootnote{\xdef\@thefnmark{}\@footnotetext}
\makeatother

\usepackage[
backend=biber,
style=ieee,
citestyle=numeric-comp,
maxbibnames=3,
maxcitenames=3,
doi=false,isbn=false,url=false,eprint=false
]{biblatex}

\defbibheading{bibliography}[\refname]{}

\title{Joint Modeling of Code-Switched and Monolingual ASR\\via Conditional Factorization}
\name{
\begin{tabular}{c}
\it Brian Yan${}^1$, Chunlei Zhang${}^2$, Meng Yu${}^2$, Shi-Xiong Zhang${}^2$, Siddharth Dalmia${}^1$, Dan Berrebbi${}^1$, \\
\it Chao Weng${}^3$, Shinji Watanabe${}^1$, Dong Yu${}^2$
\end{tabular}
}
\address{${}^1$Carnegie Mellon University, USA, ${}^2$Tencent AI Lab, USA, ${}^3$Tencent AI Lab, China}
\begin{document}
\ninept
\maketitle
\begin{abstract}

Conversational bilingual speech encompasses three types of utterances: two purely monolingual types and one intra-sententially code-switched type. 
In this work, we propose a general framework to jointly model the likelihoods of the monolingual and code-switch sub-tasks that comprise bilingual speech recognition.
By defining the monolingual sub-tasks with label-to-frame synchronization, our joint modeling framework can be \textit{conditionally factorized} such that the final bilingual output, which may or may not be code-switched, is obtained given only monolingual information.
We show that this conditionally factorized joint framework can be modeled by an end-to-end differentiable neural network. 
We demonstrate the efficacy of our proposed model on bilingual Mandarin-English speech recognition across both monolingual and code-switched corpora.

\end{abstract}
\begin{keywords}
code-switched ASR, bilingual ASR, RNN-T
\end{keywords}
\vspace{-3pt}
\section{Introduction}
\vspace{-3pt}
\label{sec:intro}

Conversational spoken language defies monolithic form, but rather is highly adaptive to situational cues such as who you are speaking to or what you are speaking about \cite{auer2013code, dragojevic2015communication}.
For instance, bilingual speakers often code-switch between different languages
to facilitate communication \cite{heredia2001bilingual, hou2020study}.
In fact, the act of bilingual code-switching itself may convey various aspects about the speaker such as a desire to spark an interpersonal connection \cite{ahn2020code}, a level of subject-matter expertise \cite{yoder2017code}, or an affinity to some socio-economic status \cite{liu2019attitudes}.\blfootnote{Work done while Brian was interning with Tencent AI Lab.}

In order to broadly cover bilingual speech, recognition systems need to recognize not only monolingual utterances from two different languages, but also intra-sententially code-switched (CS) utterances where both languages are present \cite{heredia2001bilingual, hou2020study}.
While recent advancements in the related field of multilingual speech recognition have significantly improved the language coverage of a single system by training on mixtures of monolingual utterances \cite{adams2019massively, pratap2020massively}, these works typically do not account for intra-sentential CS.
Prior works have adapted large-scale multilingual models to more flexibly identify language switch points \cite{li2019bytes, Zhou2021ACM}, 
but performance is dependent on the cross-lingual dynamics of the selected languages \cite{seki2018end}.

Another approach is to directly optimize towards intra-sentential CS for a particular bilingual pair.
A significant portion of recent works are ameliorating the linguistic differences between two unrelated languages by explicitly defining cross-lingual phone-merging rules \cite{lyudovyk2014code, luo2018towards, sivasankaran2018phone} or by implicitly learning latent language identity representations \cite{yilmaz2016investigating, song2017investigating, zeng2018end, kim2018towards, shan2019investigating, Zhang2021RnntransducerWL}.
The other significant portion of recent works are ameliorating the scarcity of paired CS speech data through data efficient methods that incorporate monolingual data into both acoustic \cite{lu2020bi, zhou2020multi, dalmia2021transformer, zhang2021decoupling} and language modeling \cite{gonen2018language, shan2019component, chuang2020training, dalmia2021transformer, zhang2021decoupling} as well as through data augmentation techniques that generate synthetic CS data \cite{chang2018code, pratapa2018language, ma2019comparison, lee2019linguistically}.
However, works focusing narrowly on intra-sentential CS often ignore or sacrfice performance on monolingual scenarios \cite{shah2020learning, sitaram2019survey}

We are interested conversational bilingual speech recognition systems (ASR) that can cover both monolingual and intra-sententially CS scenarios.
In particular, we are interested in systems that can 1) indiscriminately recognize both monolingual and intra-sententially CS utterances, 
2) efficiently leverage monolingual and CS ASR training data, and
3) be built in an end-to-end manner. %that avoids hand-crafted features.

We first propose a formulation of the bilingual ASR problem as a \textit{conditionally factorized} joint model of monolingual and CS ASR where the final output is obtained given only monolingual label-to-frame synchronized information \Sref{sec:formulation}. 
We then apply an end-to-end differentiable neural network, which we call the Conditional RNN-Transducer (RNN-T), to model our conditional joint formulation \Sref{sec:cond-rnnt}.
We show the efficacy of the Conditional RNN-T model in both monolingual and CS scenarios compared to several baselines \Sref{sec:results}.
Next, we demonstrate the Language-Separation ability, an added benefit of our proposed model \Sref{sec:separation}.
Finally, we validate a key conditional independence assumption in our framework by showing experimentally that given monolingual label-to-frame information, no other information from the observation is required \Sref{sec:3enc-ablation}.

\begin{figure}[t]
\centering
    \begin{tikzpicture}[>=stealth, node distance=2mm and 8mm]% 
    \node[circle, draw] (1) {\fontsize{8}{8}$Y$};
    \node[circle, draw] (2) [below= 1.8cm of 1] {\fontsize{8}{8}$X$};
    \draw[->] (2) -- (1);
    \node (2) [above=0mm of 1] {\textsc{Single}};
    
    \node[circle, draw, right=0.6cm of 1, inner sep=1pt] (4) {\fontsize{8}{8}$Y,I$};
    \node[circle, draw] (5) [below= 1.8cm of 4] {\fontsize{8}{8}$X$};
    \draw[->] (5) -- (4);
    \node (6) [above=0mm of 4] {\textsc{Multi}};
    
    \draw[dashed] (1.85, -2.8) -- (1.85,.7);
    \node (a) [below left=-.3cm and 0.1cm of 5] {(a)};
    
    \node[circle, draw, right= 1cm of 4] (7) {\fontsize{8}{8}$Y$};
    \node[circle, draw] (8) [below= 1.8cm of 7] {\fontsize{8}{8}$X$};
    % \node[circle, draw] (9) [above= 0.4cm of 8] {$\mathbf{l}$};
    \node[circle, draw] (9) [above= 0.4cm of 8, inner sep=3.8pt] {\fontsize{8}{8}$I$};
    % \node[circle, draw] (10) [above left= 0.4cm and 0.1cm of 9, inner sep=0.0pt] {$\mathbf{y}^M$};
    % \node[circle, draw] (11) [above right= 0.4cm and 0.1cm of 9, inner sep=0.7pt] {$\mathbf{y}^E$};
    \node[circle, draw] (10) [above left= 0.3cm and 0.1cm of 9, inner sep=.5pt] {\fontsize{8}{8}$Y^M$};
    \node[circle, draw] (11) [above right= 0.3cm and 0.1cm of 9, inner sep=1pt] {\fontsize{8}{8}$Y^E$};
    \draw[->] (8) -- (9);
    \draw[->] (9) -- (10);
    \draw[->] (9) -- (11);
    \draw[->] (10) -- (7);
    \draw[->] (11) -- (7);
    \draw[->] (9) -- (7);
    \draw[->] (8)  -- ($(10)-(0.1,0.3)$);
    \draw[->] (8) -- ($(11)-(0-.1,0.3)$);
    \node (12) [above=0mm of 7] {\textsc{Separation}};
    
    \node[circle, draw, right= 1.3cm of 7] (13) {\fontsize{8}{8}$Y$};
    \node[circle, draw] (14) [below= 1.8cm of 13] {\fontsize{8}{8}$X$};
    % \node[circle, draw] (15) [above left= 0.4cm and 0.1cm of 14, inner sep=0.0pt] {$\mathbf{y}^M$};
    % \node[circle, draw] (16) [above right= 0.4cm and 0.1cm of 14, inner sep=0.7pt] {$\mathbf{y}^E$};
    \node[circle, draw] (15) [above left= 0.4cm and 0.1cm of 14, inner sep=.5pt] {\fontsize{8}{8}$Y^M$};
    \node[circle, draw] (16) [above right= 0.4cm and 0.1cm of 14, inner sep=1pt] {\fontsize{8}{8}$Y^E$};
    % \node[circle, draw] (17) [above= 1cm of 14] {$\mathbf{g}$};
    \node[circle, draw] (17) [above= 1cm of 14] {\fontsize{8}{8}$G$};
    \draw[->] (14) -- (15);
    \draw[->] (14) -- (16);
    \draw[->] (15) -- (17);
    \draw[->] (16) -- (17);
    \draw[->] (17) -- (13);
    \draw[->] (14) -- (17);
    \draw[->] ($(15)-(0.1,-0.3)$) -- (13);
    \draw[->] ($(16)-(-0.1,-0.3)$) -- (13);
    \node (12) [above=0mm of 13] {\textsc{Mixture}};
    
    \draw[dashed] (5.75, -2.8) -- (5.75,.7);
    \node (b) [right=2.6cm of a] {(b)};
    
    \node[circle, draw, right= 1.45cm of 13] (18) {\fontsize{8}{8}$Z$};
    \node[circle, draw] (19) [below= 1.8cm of 18] {\fontsize{8}{8}$X$};
    \node[circle, draw] (20) [above left= 1cm and 0.1cm of 19, inner sep=0.5pt] {\fontsize{8}{8}$Z^M$};
    \node[circle, draw] (21) [above right= 1cm and 0.1cm of 19, inner sep=1pt] {\fontsize{8}{8}$Z^E$};
    \draw[->] (19) -- (20);
    \draw[->] (19) -- (21);
    \draw[->] (20) -- (18);
    \draw[->] (21) -- (18);
    \node (12) [above=0mm of 18] {\textsc{Conditional}};
    
    \node [right=1.8cm of b] {(c)};

\end{tikzpicture}%
    \vspace{-20pt}
    \caption{Probabilistic graphical model representations of different formulations of the bilingual ASR task, separated into three categories: (a) direct, (b) divide-and-conquer, and (c) conditional.}
    \label{fig:graphs}
    \vspace{-10pt}
\end{figure}
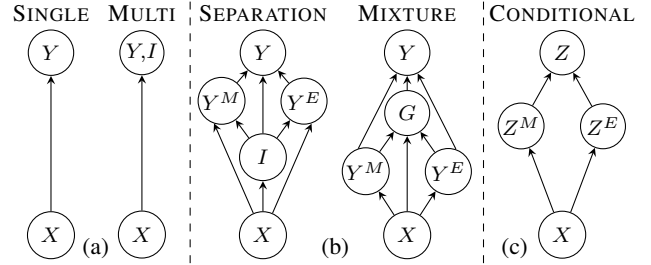

\vspace{-3pt}
\section{Background and Motivation}
\vspace{-3pt}
\label{sec:background}
In this section, we interpret the underlying probabilistic graphical models of prior works in bilingual ASR, as applied to Mandarin-English, to motivate our conditionally factorized framework in \Sref{sec:proposed}.

\vspace{-4pt}
\subsection{Direct Graphical Models}
\vspace{-2pt}

Bilingual ASR is a sequence mapping from a $T$-length speech feature sequence, $X = \{ \mathbf{x}_t \in \mathbb{R}^D | t = 1, ..., T \}$, to an $N$-length label sequence, $Y = \{ y_t \in ( \mathcal{V}^M \cup \mathcal{V}^E ) | n = 1, ..., L \}$ consisting of 
Mandarin $\mathcal{V}^M$ and English $\mathcal{V}^E$.
As shown in \Fref{fig:graphs}.a, the simplest approach models the bilingual output $Y$ as a random variable with a \textit{single} dependency on the observation $X$. 
Both the Mandarin $\mathcal{V}^M$ and English $\mathcal{V}^E$ vocabularies are regarded as part of the same set of output units and no explicit distinctions are made between the languages. 
This unification of two languages from unrelated families complicates the ASR task via phonetic ambiguities, but hand-crafting phone-merging rules may alleviate this issue \cite{lyudovyk2014code, luo2018towards, sivasankaran2018phone}.

Alternatively, \textit{multi}-tasking with language-identification can induce useful latent distinctions between Mandarin and English representations.
Here, we introduce language ID $I$ as another random variable with a single dependency on the observation $X$.
Note that the bilingual output $Y$ and the language-ID $I$ do not interact with each other aside from being implicitly related as they both directly depend on $X$.
During training, this multi-tasking approach helps to resolve cross-lingual ambiguities within the shared bilingual latent representations \cite{yilmaz2016investigating, song2017investigating, zeng2018end, kim2018towards, shan2019investigating, Zhang2021RnntransducerWL}.
During inference, the bilingual output is obtained directly from the observation similar to the single dependency approach.
The main drawback of these direct approaches is the potentially high complexity of the $X\rightarrow Y$ dependency due to conflicting linguistics of different two languages. 
\vspace{-5pt}

\subsection{Divide-and-Conquer Graphical Models}
\vspace{-2pt}
Alternatively, the bilingual ASR task can be decomposed into its sub-component monolingual parts as in the divide-and-conquer approaches shown in \Fref{fig:graphs}.b.
The \textit{separational} approach uses language ID $I$ to segment the observation $X$ into monolingual parts before passing those to monolingual recognizers which separately predict Mandarin $Y^M = \{ y^M_l \in \mathcal{V}^M | l = 1, ..., L \}$ and English $Y^E = \{ y^E_l \in \mathcal{V}^E | l = 1, ..., L \}$.
Finally, the monolingual outputs $Y^M$ and $Y^E$ are stitched together accordingly to obtain the final bilingual output $Y$. 
This approach successfully introduces simpler $X\rightarrow Y^M$ and $X\rightarrow Y^E$ dependencies, but does so at the cost of the additional dependencies associated with the language ID random variable $I$: $I\rightarrow Y^M$, $I\rightarrow Y^E$, and $I\rightarrow Y$.
Therefore, performance becomes highly dependent on the ability to correctly predict the language identity at a segment level \cite{chan2004detection, weiner2012integration, lyu2013language, rallabandi2018automatic, li2019towards}.

The \textit{mixture}-based approach avoids this dependency on the language ID $I$ by instead introducing another random variable $G$, a gating mechanism.
This approach first noisily models $X\rightarrow Y^M$ and $X\rightarrow Y^E$, but maintains latent representations of each so as to avoid early decisions.
Then, it models the gating mechanism $G$ as a function of these latent monolingual representations; we approximate this dependency in our graph as $( X, Y^M, Y^E ) \rightarrow G$.
Finally, the gating mechanism $G$ is used to fuse the latent monolingual representations to ultimately obtain the bilingual output $Y$.
This is a promising approach that efficiently combines information from monolingual experts by operating entirely within the latent space \cite{lu2020bi, zhou2020multi, dalmia2021transformer}. 
However, mixture-based approaches much like their separational cousins incur the cost of the additional dependencies associated with the gating random variable $G$.
Our motivation is to decompose the bilingual ASR problem into monolingual sub-tasks without incurring any additional random variables.

\vspace{-4pt}

\section{Proposed Framework}
\vspace{-3pt}
\label{sec:proposed}
\begin{table*}[t]
  \centering
    \caption{Results comparing the Conditional RNN-T models to Vanilla and Gating RNN-T baselines on intra-sententially code-switched (CS), monolingual Mandarin, and monolingual English test sets. The upper half shows results when using only CS data during fine-tuning while the bottom half shows results when using CS + monolingual data during fine-tuning. All models are pre-trained on monolingual data.}
    \vspace{-3pt}
    % \resizebox {\linewidth} {!} {
\begin{tabular}{llcc|ccc|c|c}
\toprule
& & Pre-trained & Fine-tuning & \multicolumn{3}{c|}{\underline{\textsc{Code-Switched}}} & \underline{\textsc{Mono-Man}} & \underline{\textsc{Mono-Eng}} \\
% \cmidrule(r){4-6} \cmidrule(r){7-7} \cmidrule(r){8-8}
Model Type & Model Name & Encoder(s) & Data & MER & CER & WER & CER & WER \\
\midrule
Direct & Vanilla RNN-T \cite{Zhang2021RnntransducerWL, dalmia2021transformer} & \cmark & CS & 12.3 & 9.9 & 34.3 & 17.9 & 81.4 \\
Mixture & Gating RNN-T \cite{lu2020bi, dalmia2021transformer} & \cmark & CS & 11.5 & 9.1 & 33.0 & 17.7 & \textbf{78.3} \\
Conditional & Our Proposed Model & \cmark & CS & 11.5 & 9.1 & 33.2 & 15.5 & 82.9 \\
Conditional & + Language-Separation (LS) & \cmark & CS & \textbf{11.1} & \textbf{8.7} & \textbf{32.7} & \textbf{15.3} & 82.7 \\
\midrule
Direct & Single RNN-T \cite{Zhang2021RnntransducerWL, dalmia2021transformer} & \cmark & CS + M & 11.3 & 9.3 & 30.8 & 6.5 & 17.8 \\
Mixture & Gating RNN-T \cite{lu2020bi, dalmia2021transformer} & \cmark & CS + M & 11.2 & 8.8 & 34.7 & 5.7 & 34.6 \\
Conditional & Our Proposed Model & \cmark & CS + M & 10.3 & 8.2 & 29.5 & 5.4 & 16.5 \\
Conditional & + Language-Separation (LS) & \cmark & CS + M & \textbf{10.2} & \textbf{8.1} & \textbf{29.2} & \textbf{5.3} & \textbf{16.3} \\
\bottomrule
\end{tabular}
% }

% \resizebox {\linewidth} {!} {
% \begin{tabular}{lccccc}
% \toprule
% & \multicolumn{3}{c}{Code-Switched} & Man-Only & Eng-Only \\
% \cmidrule(r){2-4} \cmidrule(r){5-5} \cmidrule(r){6-6}
% Model & MER & CER & WER & CER & WER \\
% \midrule
% Vanilla RNN-T & 11.4 & 9.4 & 29.5 & 4.7 & 20.0 \\
% Gating RNN-T & 11.2 & 8.9 & 33.3 & 3.5 & 44.6\\
% \midrule
% Conditional RNN-T & 10.4 & 8.4 & 29.0 & 3.6 & 18.6 \\
% + Language-Masking & \textbf{10.2} & \textbf{8.2} & \textbf{28.0} & \textbf{3.3} & \textbf{18.4} \\
% \bottomrule
% \end{tabular}
% }
    \vspace{-4pt}
  \label{tab:main}
\end{table*}

In this section, we first propose a framework using label-to-frame synchronization to obtain bilingual outputs given only conditional monolingual information.
Then we show an end-to-end differentiable method to model our conditionally factorized framework.

\vspace{-5pt}
\subsection{Conditionally Factorized Formulation of Bilingual ASR}
\label{sec:formulation}

Rather than treating CS ASR as a single sequence transduction task, we seek to decompose it into three portions: 1) recognizing Mandarin 2) recognizing English and 3) composing recognized monolingual segments into a bilingual sequence which may or may not be CS.
However, given only monolingual portions of the output $Y^M$ and $Y^E$ we cannot form $Y$ without the order in which they should be composed.
Therefore, in order to satisfy our desired \textit{conditional} probabilistic graph (shown in \Fref{fig:graphs}.c), we need richer monolingual representations which contain  ordering information.

Consider that for each output $T$-length observation sequence $X$ and $L$-length bilingual label sequence $Y$, there are a number of possible $T$-length label-to-frame sequences $Z = \{ z_t \in \mathcal{V}^M \cup \mathcal{V}^E \cup \{\varnothing \} | t=1\ldots T \}$.
Here, $z_t$ is either a surface-level unit or a blank symbol denoting a null emission as in Connectionist Temporal Classification (CTC) \cite{graves2006connectionist}.
The posterior of the bilingual label sequence $p(Y|X)$ is then factorized as follows:
\begin{equation}
    p(Y | X) = \sum_{Z \in \mathcal{Z}(Y)} p(Z | X) \label{eq:1}
\end{equation}
Note that the summation is over all possible label-to-frame alignments $Z \in \mathcal{Z}(Y)$ for a given observation $X$ and label $Y$ pair.\footnote{$Z$ maps to $Y$ deterministically via repeat and blank removal rules \cite{rnnt_graves}.}

Next, we re-formulate any bilingual label-to-frame sequence in terms of its constituent monolingual label-to-frame sequences $Z^M = \{ z^M_t \in \mathcal{V}^M \cup \{\varnothing \} | t=1\ldots T \}$ and $Z^E = \{ z^E_t \in \mathcal{V}^E \cup \{\varnothing \} | t=1\ldots T \}$.
Now, we can indeed obtain the bilingual $Z$ given only conditional monolingual information $Z^M$ and $Z^E$:
\begin{equation}
    z_t = 
    \begin{cases}
        z^M_t, & \text{if $z^M_t \in \mathcal{V}^M$ and $z^E_t = \varnothing$} \\
        z^E_t, & \text{if $z^E_t \in \mathcal{V}^E$ and $z^M_t = \varnothing$} \\
        \varnothing, & \text{if $z^M_t = \varnothing$ and $z^E_t = \varnothing$} \label{eq:2}
    \end{cases}
\end{equation}
Note that at position $t$ only one of $z^M_t$ or $z^E_t$ may be a surface-level unit while the other must be the blank symbol by definition. It is possible that both $z^M_t$ and $z^E_t$ are blank symbols, but it is not possible that both are surface-level units.

Following this interpretation of the bilingual sequence $Z$ in terms of its monolingual parts $Z^M$ and $Z^E$, we can re-formulate the posterior $p(Z|X)$ as a joint likelihood $p(Z, Z^M, Z^E | X)$:
\begin{align}
    p(Z | X) &= p(Z, Z^M, Z^E | X) \\
    &=  p(Z | Z^M, Z^E, X)  p(Z^M, Z^E | X) \label{eq:halfstep} \\
    &\approx p(Z | Z^M, Z^E, \cancel{X})p(Z^M|X)p(Z^E|X) \label{eq:4}
\end{align}
From Eq. \eqref{eq:halfstep} to Eq. \eqref{eq:4}, we make two key independence assumptions.
The first is that given $Z^M$ and $Z^E$, no other information from the observation $X$ is required to determine $Z$ since the monolingual label-to-frame sequences already contain ordering information.
We show that this assumption holds experimentally in \Sref{sec:3enc-ablation}.
The second assumption is that $p(Z^M|X)$ and $p(Z^E|X)$ are independent, allowing for separate modeling of monolingual posteriors.

\vspace{-3pt}
\subsection{Conditional RNN Transducer}
\label{sec:cond-rnnt}

\subsubsection{Monolingual Modules}
The monolingual label-to-frame posteriors $p(Z^M|X)$ and $p(Z^E|X)$ are further factorized with the chain-rule as follows:
\vspace{-2pt}
\begin{align}
    p(Z^M|X) &\approx \prod_{t=1}^{T} p(z_t^M | X, \cancel{z^M_{1:t-1}}) \label{eq:5} \\
    p(Z^E|X) &\approx \prod_{t=1}^{T} p(z_t^E | X, \cancel{z^E_{1:t-1}}) \label{eq:6}
\end{align}
Note that we make the conditional independence assumption here as a modeling choice, mitigating label / exposure bias issues \cite{ranzato2015sequence, bottou1997global}.

Our proposed Conditional RNN-T models the posterior as follows.
Given the observed speech feature sequence $X$, a Mandarin-only $\textsc{Encoder}_{M}$ maps to a sequence of hidden representations $\mathbf{h}^{M} = \{ \mathbf{h}_t^M \in \mathbb{R}^{D} | t = 1,  ... , T\}$ and an English-only $\textsc{Encoder}_{E}$ maps to a sequence of hidden representations $\mathbf{h}^{E} = \{ \mathbf{h}_t^E \in \mathbb{R}^{D} | t = 1,  ... , T\}$.
Separate linear projection layers followed by softmax activations, $\textsc{SoftmaxOut}_M$ and $\textsc{SoftmaxOut}_M$, yield the posteriors $p(z_t^M | X)$ and $p(z_t^E | X)$.
We train these modules using CTC loss functions $\mathcal{L}_\text{M\_CTC}$ and $\mathcal{L}_\text{E\_CTC}$ \cite{graves2006connectionist}.

\vspace{-3pt}
\subsubsection{Bilingual Module}
\vspace{-3pt}
The bilingual conditional likelihood $p(Z | Z^M, Z^E)$ is further factorized with the chain-rule as well:
\begin{align}
    p(Z | Z^M, Z^E) = \prod_{i=1}^{T+L} p(z_i | Z^M, Z^E, z_{1:i-1}) \label{eq:11}
\end{align}

The monolingual alignment information is passed to the bilingual module via the hidden representations $\mathbf{h}^M$ and $\mathbf{h}^E$. We therefore approximate $P(z_i|Z^M, Z^E, z_{1:i-1})$ as follows:
\begin{align}
    \mathbf{h}^{\textsc{Enc}}_t &= \mathbf{h}_t^M + \mathbf{h}_t^E \label{eq:fuse} \\
    \mathbf{h}^{\textsc{Dec}}_l &= \textsc{Decoder}(z_{1:l-1}) \\
    \mathbf{h}^{\textsc{Jnt}}_{t,l} &= \textsc{Joint}(\mathbf{h}^{\textsc{Enc}}_t, \mathbf{h}^{\textsc{Dec}}_l) \\
    p(z_i | \mathbf{h}^M, \mathbf{h}^E, z_{1:i-1}) &= \textsc{SoftmaxOut}(\mathbf{h}^{\textsc{Jnt}}_{t,l}) \label{eq:15}
\end{align}
Note that \eqref{eq:15} approximates $p(z_i|Z^M, Z^E, z_{1:i-1})$ from \eqref{eq:11} by using the latent representations $\mathbf{h}^M$ and $\mathbf{h}^E$ to pass the monolingual label-to-frame information $Z^M$ and $Z^E$.\footnote{Alternatively, logit or softmax normalized could be used here \cite{dalmia2019enforcing}.}
We train these modules using the RNN-T loss function $\mathcal{L}_\text{RNNT}$ \cite{rnnt_graves}.

\vspace{-3pt}
\subsubsection{Full Network}
\vspace{-3pt}
\label{sec:training}
With Equations \eqref{eq:1} and \eqref{eq:4}, the posterior $p(Y|X)$ finally becomes:% represented as follows:
\begin{align}
    p(Y | X) &\approx \underbrace{\sum_{\mathcal{Z}} p(Z | Z^M, Z^E)}_{\overset{\Delta}{=} p_{\text{rnnt}}(Y | Z^M, Z^E)} 
        \underbrace{\sum_{\mathcal{Z}^M} p(Z^M|X)}_{\overset{\Delta}{=} p_{\text{ctc}}(Y^M | X)} 
        \underbrace{\sum_{\mathcal{Z}^E} p(Z^E|X)}_{\overset{\Delta}{=} p_{\text{ctc}}(Y^E | X)} \label{eq:full}
\end{align}
where the monolingual CTC, $p_{\text{ctc}}(Y^M | X)$ and $p_{\text{ctc}}(Y^E | X)$, and bilingual RNN-T, $p_{\text{rnnt}}(Y | Z^M, Z^E)$, objective functions are defined as summations over all possible frame-to-label sequences $Z^M \in \mathcal{Z}^M(Y^M)$, $Z^E \in \mathcal{Z}^E(Y^E)$, and $Z \in \mathcal{Z}(Y)$ respectively.

We train our Conditional RNN-T model using an initial monolingual pre-training step to maximize $p_{\text{ctc}}(Y^M | X)$ and $p_{\text{ctc}}(Y^E | X)$. We then fine-tune the entire network to maximize $p_{\text{rnnt}}(Y | Z^M, Z^E)$, including both the bilingual and pre-trained monolingual modules, on a mixture of monolingual and CS data. Since this first regime only implicitly applies monolingual conditioning via pre-trained initialization, we propose an explicit alternative regime which we call the Conditional RNN-T + Language-Separation (LS). This variant of our proposed model uses a multi-tasked loss $\mathcal{L}_{LS}$ with tunable $\lambda$:
\begin{align}
    \mathcal{L}_\text{LS} = \lambda \mathcal{L}_\text{RNNT} + (1 - \lambda) (\mathcal{L}_\text{M\_CTC} + \mathcal{L}_\text{E\_CTC}) \label{eq:LS}
\end{align}
The monolingual ground truths $Y^M$ and $Y^E$ are obtained by applying a language-specific mask to the bilingual ground truth $Y$.\footnote{This masking is applicable to both monolingual and CS $Y$. E.g. if $y_l \in \mathcal{V}^M \forall l$, then the masked $Y^E$ is the empty string and $Y^M$ is the entire $Y$.}

\section{Data and Experimental Setup}
\vspace{-3pt}
\label{sec:setup}

\mypar{Data:} We use $200$h of intra-sententially CS training data from the ASRU 2019 shared task where Mandarin is the matrix and English is the embedded language \cite{shi2020asru}. 
We use $500$h of monolingual Mandarin data for pre-training and a $200$h subset for fine-tuning \cite{shi2020asru}.
We use $700$h of accented monolingual English data for pre-training and a $200$h subset for fine-tuning \cite{k190}.
We use provided test CS data and generate our own splits for monolingual test sets.

\mypar{Experimental Setup:} All our models were trained using the ESPnet toolkit \cite{watanabe2018espnet}.
Our input features are global mean-variance normalized 83 log-mel filterbank and pitch features \cite{povey2011kaldi}.
We apply the Switchboard Strong (SS) augmentation policy of SpecAugment \cite{park2019specaugment}.
We combine $5000$ Mandarin characters with 5000 English BPE \cite{sennrich2015neural} units to form the output vocabulary.
$\textsc{Encoders}$ are conformers \cite{gulati2020conformer, guo2021recent} with $12$ blocks, kernel size of $15$, $2048$ feed-forward dim, $256$ attention dim, and 4 heads.
$\textsc{Decoders}$ are LSTMs \cite{hochreiter1997long, watanabe2018espnet} with $1$ layer, $1024$ embed dim, $512$ unit dim, and $512$ joint dim.
The direct RNN-T baseline with only $1$ encoder uses a doubled $512$ attention dim, so all models have about $100$M params. 
We use the Adam optimizer to train 80 epochs with an inverse square root decay schedule, a transformer-lr scale~\cite{watanabe2018espnet} of $1$, $25$k warmup steps, and an effective batchsize of $192$.
$\textsc{Encoders}$ are pretrained using the hybrid CTC/Attention framework \cite{watanabe2017hybrid}.
We use beam-size of 10 during inference. Ablation studies on CTC sub-nets use greedy decoding.

\section{Results}
\vspace{-3pt}
\label{sec:results}

In \Tref{tab:main}, we compare the CS and monolingual performances of our Conditional RNN-T and Conditional RNN-T + LS models to direct and mixture-based baselines, which are our re-implementations of Vanilla RNN-T \cite{dalmia2021transformer, Zhang2021RnntransducerWL} and Gating RNN-T \cite{lu2020bi, dalmia2021transformer} described in prior works.
The top and bottom portions of \Tref{tab:main} compare results when using only CS data versus using both CS and monolingual data during fine-tuning.
Not only did all models perform significantly better on monolingual sets when using monolingual fine-tuning data, they also improved on the CS set suggesting that the monolingual data is indeed supplementing the CS training data.\footnote{Unlike the shared task in \cite{shi2020asru}, we evaluate on monolingual corpora and do not use the provided 3-gram LM, the full monolingual data during fine-tuning, data augmentation besides SpecAugment, or LID multi-tasking.}

As shown in the bottom half of \Tref{tab:main}, the Gating RNN-T slightly outperforms the Vanilla RNN-T on CS and monolingual Mandarin but is degraded on monolingual English, suggesting that the gating mechanism overly focuses on outputting Mandarin due to the high skew of the CS training data towards Mandarin. 
Our proposed Conditional RNN-T model outperforms the best baselines consistently across evaluation sets.
On the monolingual sets, the Conditional RNN-T model performs similarly to monolingual-only models trained on the same data.
Further, the Language-Separation loss incrementally improves across all sets suggesting a benefit to the explicit monolingual conditioning method described in Eq. \eqref{eq:LS}. We examine this benefit further in the subsequent sub-section.

\subsection{Language-Separation Ability of Monolingual Modules}
\label{sec:separation}

\begin{figure}[t]
  \centering
    \includegraphics[width=0.85\linewidth]{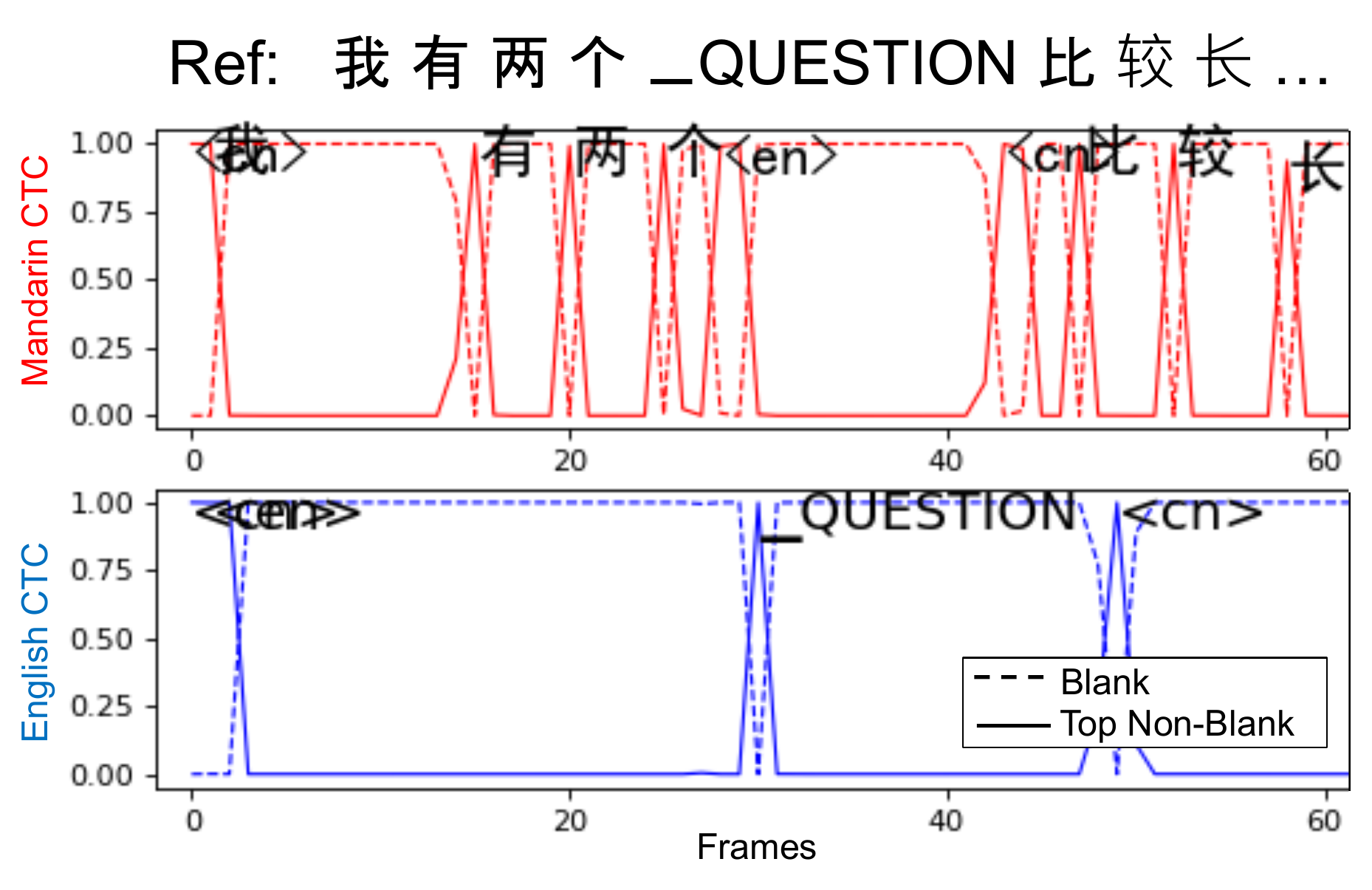}
    \vspace{-15pt}
    \caption{Example illustrating the label-to-frame posteriors of the monolingual CTC sub-nets for an intra-sententially CS utterance.}
    % \vspace{-15pt}
    \label{fig:segmentation}
\end{figure}

\begin{table}[t]
  \centering
    \caption{Ablation study examining the language-separation ability of the monolingual CTC sub-nets, $p(Z^M|X)$ and $p(Z^E|X)$, on the CS dev set. The sub-nets are expected to transcribe speech from their languages while ignoring speech in the other. Performance is evaluated using monolingual parts $Y^M$ and $Y^E$ of the ground-truth CS label sequence $Y$. CER/WER and Insertion Rate (INS) are shown.}
    \vspace{-3pt}

\resizebox {\linewidth} {!} {
\begin{tabular}{lccc|ccc}
\toprule
& \multicolumn{3}{c}{\underline{\textsc{Man Portion of CS}}} & \multicolumn{3}{c}{\underline{\textsc{Eng Portion of CS}}} \\
% \cmidrule(r){2-4} \cmidrule(r){5-5} \cmidrule(r){6-6}
Model & Sub-Net & CER & INS & Sub-Net & WER & INS \\
\midrule
% \multirow{2}{*}{Gating RNN-T} & $P(\mathbf{z}^M | \mathbf{x})$ & 14.0 & - & 4.1 & - \\
% & $P(\mathbf{z}^E | \mathbf{x})$ & - & 318.7 & - & 99.7 \\
% \midrule
Cond. RNN-T & $p(Z^M | X)$ & 11.8 & 3.7 & $p(Z^E | X)$ & 42.7 & 7.9 \\
Cond. RNN-T + LS & $p(Z^M | X)$ & \textbf{8.6} & \textbf{0.7} & $p(Z^E | X)$ & \textbf{37.1} & \textbf{4.6}\\
% \midrule
% % Cond RNN-T + LS & - & \textbf{8.2} & \textbf{28.0} & \textbf{3.3} & \textbf{18.4} \\
% \multirow{2}{*}{Cond RNN-T + LS} & $P(\mathbf{z}^M | \mathbf{x})$ & \textbf{8.6} & - \\
% & $P(\mathbf{z}^E | \mathbf{x})$ & - & \textbf{37.1} \\
% \midrule
% Mandarin CTC & N/A & X & - & X & - \\
% English CTC & N/A & - & X & - & X \\
\bottomrule
\end{tabular}
}

% 31.0 3.8 7.9
% 28.9 3.5 4.6

% 7.8 0.3 3.7
% 7.4 0.5 0.7
  \label{tab:subnet}
\end{table}

Recall that in Eq. \eqref{eq:2} we state that the bilingual $Z$ is specified in terms of its monolingual parts, $Z^M$ and $Z^E$, such that at any position $t$ only one of $z^M_t$ or $z^E_t$ may be non-blank.
If our proposed Conditional RNN-T model is indeed modeling $p(Z^M | X)$ and $p(Z^E | X)$ with this property, it logically follows that each monolingual CTC sub-net would be capable of emitting labels for frames corresponding to their language while emitting blanks for frames in the other language; we refer to this diarization-like \cite{lyu2013language} ability as Language-Separation.
This Language-Separation is observable in \Fref{fig:segmentation} which depicts the blank and non-blank posterior values of each monolingual CTC sub-net for a snippet of CS speech.
Here, the Mandarin side emits characters while the English side emits blanks except for frames $30$ to $40$ where the opposite occurs.

As mentioned in \Sref{sec:training}, there are two ways to optimize the Conditional RNN-T model towards the conditionally factorized formulation in Eq. \eqref{eq:full}.
\Tref{tab:subnet} shows that the implicit way, which uses pre-training to condition the bilingual task $p(Z|Z^M, Z^E)$, produces monolingual CTC sub-nets that reasonably able perform Language-Separation on CS data. 
However, the explicit way described in Eq. \eqref{eq:LS} is preferred since it provides a supervised Language-Separation signal.
Thus, the resultant monolingual CTC sub-nets of this Conditional RNN-T + LS model have a greater Language-Separation ability as indicated by the reduced insertion errors for the same CS data.

\subsection{Conditional Independence of $p(Z | Z^M, Z^E)$ from $X$}
\label{sec:3enc-ablation}

\begin{table}[t]
  \centering
    \vspace{-3pt}
    \caption{Experimental validation of conditional independence assumption in the bilingual module which models $p(Z | Z^M, Z^E, \cancel{X})$. The 3-Encoder variant removes this assumption in its bilingual module $p(Z | Z^M, Z^E, X)$. Results are shown on the CS dev set.}
    \vspace{-3pt}
    % \resizebox {\linewidth} {!} {
% \begin{tabular}{lc|ccc}
% \toprule
% & & \multicolumn{3}{c}{\underline{\textsc{Code-Switch}}}\\
% % \cmidrule(r){2-4} \cmidrule(r){5-5} \cmidrule(r){6-6}
% Model & Sub-Net & MER & CER & WER \\
% \midrule
% \multirow{3}{*}{Cond RNN-T + LS} & $P(\mathbf{z}^M | \mathbf{x})$ & - & 8.8 & - \\
% & $P(\mathbf{z}^E | \mathbf{x})$ & - & - & 36.6 \\
% & $P(\mathbf{z} | \mathbf{z}^M, \mathbf{z}^E)$ & \textbf{11.1} & 8.9 & \textbf{31.1} \\
% \midrule
% % Cond RNN-T + LS & - & \textbf{8.2} & \textbf{28.0} & \textbf{3.3} & \textbf{18.4} \\
% \multirow{4}{*}{3-Enc RNN-T + LS} & $P(\mathbf{z}^M | \mathbf{x})$ & - & \textbf{8.7} & - \\
% & $P(\mathbf{z}^E | \mathbf{x})$ & - & - & 36.1 \\
% & $P(\mathbf{z} | \mathbf{x})$ & 100.0 & 100.0 & 100.0 \\
% & $P(\mathbf{z} | \mathbf{z}^M, \mathbf{z}^E, \mathbf{x})$ & 11.2 & 9.0 & \textbf{31.1} \\
% \bottomrule
% \end{tabular}
% }

\resizebox {\linewidth} {!} {
\begin{tabular}{lc|ccc}
\toprule
& Bilingual & \multicolumn{3}{c}{\underline{\textsc{Code-Switched}}}\\
% \cmidrule(r){2-4} \cmidrule(r){5-5} \cmidrule(r){6-6}
Model & Condition & MER & CER & WER \\
\midrule
% \multirow{3}{*}{Cond RNN-T + LS} & $P(\mathbf{z}^M | \mathbf{x})$ & - & 8.8 & - \\
% & $P(\mathbf{z}^E | \mathbf{x})$ & - & - & 36.6 \\
Cond. RNN-T + LS & $p(Z | Z^M, Z^E)$ & \textbf{11.1} & \textbf{8.9} & \textbf{31.1} \\
% \midrule
% Cond RNN-T + LS & - & \textbf{8.2} & \textbf{28.0} & \textbf{3.3} & \textbf{18.4} \\
% \multirow{4}{*}{3-Enc RNN-T + LS} & $P(\mathbf{z}^M | \mathbf{x})$ & - & \textbf{8.7} & - \\
% & $P(\mathbf{z}^E | \mathbf{x})$ & - & - & 36.1 \\
% & $P(\mathbf{z} | \mathbf{x})$ & 100.0 & 100.0 & 100.0 \\
3-Enc. RNN-T + LS & $p(Z | Z^M, Z^E, X)$ & 11.2 & 9.0 & \textbf{31.1} \\
\bottomrule
\end{tabular}
}
  \label{tab:3-encoder}
\end{table}

Finally, we experimentally validate the conditional independence assumption in Eq. \eqref{eq:4} that the final bilingual output $Z$ depends only on monolingual alignment information $Z^M$ and $Z^E$ and nothing else from the observation $X$.
In this study, we augment the Conditional RNN-T + LS model with a third $\textsc{Encoder}_A$ which maps the observation $X$ to hidden representations $\mathbf{h}^{A} = \{ \mathbf{h}_t^A \in \mathbb{R}^{D} | t = 1,  ... , T\}$.
This $\mathbf{h}^A$ is then added as a third term to the fusion in Eq. \eqref{eq:fuse} thereby allowing the bilingual module to model $p(Z | Z^M, Z^E, X)$ instead of $p(Z | Z^M, Z^E)$.
We call this modified model the 3-Encoder RNN-T + LS and train it in the same way as the Conditional RNN-T + LS model.
As shown in \Tref{tab:3-encoder}, this 3-Encoder variant does not capture any additional useful information from the observation $X$ but rather performs slightly worse than our Conditional RNN-T + LS model.

\vspace{-3pt}
\section{Conclusion}
\vspace{-3pt}
We present an end-to-end framework for jointly modeling CS and monolingual ASR with conditional factorization such that the bilingual task is logically decomposed into simpler sub-components. 
We show improvements on both CS and monolingual ASR over prior works, suggesting that our general joint modeling approach is promising towards building robust bilingual systems.
In future work, we seek to extend our approach with joint decoding \cite{watanabe2017hybrid}, differentiable WFST representations of code-switching linguistics \cite{yan2021differentiable}, and modular neural network methods \cite{dalmia2019enforcing}.

\vspace{-3pt}
\section{Acknowledgements}
\vspace{-3pt}
We would like to thank Florian Boyer, Jia Cui, Xuankai Chang, Hirofumi Inaguma, and Koshak for their support.

\vfill\pagebreak


\section{References}
{
\printbibliography
}

\end{document}